\documentclass[lettersize,journal]{IEEEtran}
\usepackage{amsmath,amsfonts}
\usepackage{algorithm}
\usepackage{algorithmic}
\usepackage{array}
\usepackage[caption=false,font=normalsize,labelfont=sf,textfont=sf]{subfig}
\usepackage{textcomp}
\usepackage{stfloats}
\usepackage{url}
\usepackage{verbatim}
\usepackage{cite}
\usepackage{color}

\usepackage{xcolor}
\definecolor{citecolor}{HTML}{0071bc}
\usepackage[colorlinks, linkcolor=red,  anchorcolor=blue, citecolor=citecolor]{hyperref}

\usepackage{multirow}
\usepackage{graphicx}
\def\BibTeX{{\rm B\kern-.05em{\sc i\kern-.025em b}\kern-.08em
    T\kern-.1667em\lower.7ex\hbox{E}\kern-.125emX}}
\usepackage{balance}
\begin{document}
\title{AMatFormer: Efficient Feature Matching via Anchor Matching Transformer}
\author{Bo Jiang, Shuxian Luo, Xiao Wang*, \emph{Member, IEEE}, Chuanfu Li and Jin Tang  
\thanks{
Bo Jiang is with the
 Information Materials and Intelligent Sensing Laboratory of Anhui Province, School of Computer Science and Technology, Anhui University, Hefei 230601, China. 
Shuxian Luo, Xiao Wang and Jin Tang are with the School of Computer Science and Technology, Anhui University, Hefei 230601, China. 
Chuanfu Li is with the First Affiliated Hospital of Anhui University of Chinese Medicine, Hefei, China.
}
\thanks{Email: jiangbo@ahu.edu.cn, 17305693483@163.com, wangxiaocvpr@foxmail.com, tangjin@ahu.edu.cn} 
\thanks{* Corresponding author: Xiao Wang} 
}

\markboth{Journal of \LaTeX\ Class Files,~Vol., No., 2022}%
{How to Use the IEEEtran \LaTeX \ Templates}

\maketitle

\begin{abstract}
Learning based feature matching methods have been commonly studied in recent years. The core issue for learning feature matching is to how to learn (1) discriminative representations for feature points (or regions) within each intra-image and (2) consensus representations for feature points across inter-images. Recently, self- and cross-attention models have been exploited to address this issue. However, in many scenes, features are coming with large-scale, redundant and outliers contaminated. Previous self-/cross-attention models generally conduct message passing on all primal features which thus lead to redundant learning and high computational cost. To mitigate limitations, inspired by recent seed matching methods, in this paper, we propose a novel efficient Anchor Matching Transformer (AMatFormer) for the feature matching problem. AMatFormer has two main aspects: First, it mainly conducts self-/cross-attention on some anchor features and leverages these anchor features as message bottleneck to learn the representations for all primal features. Thus, it can be implemented efficiently and compactly. Second, AMatFormer adopts a shared FFN module to further embed the features of two images into the common domain and thus learn the consensus feature representations for the matching problem. Experiments on several benchmarks demonstrate the effectiveness and efficiency of the proposed AMatFormer matching approach. 
\end{abstract}


\section{Introduction}
\IEEEPARstart{F}{eature} matching which aims to establish the correspondences between features (\emph{e.g.}, points, regions) of source and target images is a fundamental   problem in computer vision and pattern recognition fields.
To conduct feature matching, it first needs to
represent each feature in both images with a discriminative descriptor.
Then, the metric learning model is employed to obtain the correlation/similarity between pairs of feature points in both images.
Finally, the linear assignment algorithm is conducted to find the correspondences between features~\cite{ma2021image}.
In recent years, deep learning models have been commonly explored for feature matching tasks~\cite{2020Learning, 2020SuperGlue, chen2021sgmnet, sun2022guide, 2015kanglearning, 2022chenmulti, 8362992}. These methods can integrate feature representation, metric learning and final correspondence estimation together in an end-to-end manner.


 The core issue for deep learning feature matching is how to learn (1) \emph{discriminative} representations for feature points within each intra-image and (2) \emph{consensus} representations for feature points across two inter-images.
 Recently, self- and cross-attention models have been exploited to address this issue. 
Specifically, self-attention is used to capture the relationships of features and thus can learn context-aware representations for feature units within each intra-image \emph{while} cross-attention is employed to model the interactions between features of two inter-images and thus can conduct information communication across inter-images.
For example,
for the graph matching problem, attention based on graph learning techniques have been incorporated into the graph matching network~\cite{liu2021joint,jiang2019glmnet}.
Sarlin et al.,~\cite{2020SuperGlue} propose SuperGlue which uses self-attention and cross-attention to leverage both spatial
relationships of the keypoints and their visual appearance for intra and inter-image feature representation.
Jiang et al.,~\cite{Jiang_2021_ICCV} propose Correspondence Transformer for Matching (COTR) which adopts the regular Transformer as the feature enhancement module for feature correspondence problem.

%

 However, in many scenes, features are coming with large-scale, redundant and outliers contaminated.
Previous self-/cross-attention models generally conduct information mixing on all primal features which thus
lead to redundant learning and high computational cost.
To address this issue,
Chen et al.,~\cite{chen2021sgmnet} propose to develop a seeded graph neural network for feature matching problems.
It first obtains seed features from primal features for both images and then leverages seed features as attention bottlenecks to
learn the representations for all primal features.
However, SGMnet~\cite{chen2021sgmnet} utilizes an attentional pooling module to obtain seed representations which is time consuming
and also redundant with the seed self-attention learning module. 
 Also, it only employs the pure self-attention~\cite{2017Attention} and message passing operation~\cite{gilmer2017neural}, lacking of feature transformation to further learn
 the consensus representations for both source and target images.

%

To mitigate these limitations, inspired by recent seed matching~\cite{fishkind2019seeded,darmon2020learning,chen2021sgmnet} and vision transformer~\cite{dosovitskiy2021an, lee2019set}, in this paper, we propose a novel efficient Anchor Matching Transformer (AMatFormer) for feature matching problem. Different from previous self-/cross-attention based methods~\cite{liu2021joint,2020SuperGlue,chen2021sgmnet}, AMatFormer adopts a Siamese Transformer architecture which involves Anchor Selection, Anchor Self/Cross Attention, Anchor-Primal Attention and Feed-Forward Network (FFN) layer, followed by some residual connections. 
The main aspects of AMatFormer are twofold. 
First, AMatFormer mainly conducts self-/cross-attention on some anchor features and leverages these anchor features as message bottleneck to learn the representations for primal features. \textcolor{black}{Comparing with SuperGlue~\cite{2020SuperGlue}, AMatFormer leverages a more efficient set Transformer~\cite{lee2019set}
for feature matching problem. 
In contrast to Seeded GNN~\cite{chen2021sgmnet}, AMatFormer employs a Transformer-type message passing mechanism~\cite{dosovitskiy2021an} and does not involve the attentional pooling module which thus performs more {efficiently} and {compactly}.}
Also, AMatFormer adopts a shared FFN module to further embed the features of two images into the common domain. Thus, it can learn the consensus feature representations for the matching problem.
%
Besides, AMatFormer employs a more flexible and compact metric learning model for feature matching tasks. 
\textcolor{black}{Thus, the proposed method can be potentially applied to many tasks, such as zero-shot temporal activity detection~\cite{zhang2022tn}, video classification, visual tracking and scene graph analysis~\cite{chang2021comprehensive}, etc.}

Overall, the main contributions of this paper are summarized as follows:
\begin{itemize}
\item We propose an efficient  Anchor Matching Transformer (AMatFormer) scheme for dual-image representations.
 It can capture the dependence of intra-image features (\emph{e.g.}, points, regions and patches) and explore the interaction between inter-images to learn consistent representations for dual images.
 \textcolor{black}{AMatFormer can be potentially used in image matching, re-identification and dual-modality learning tasks in Computer Vision and Multimedia field.}
\item We develop a novel efficient feature matching network which involves AMatFormer based feature representations, metric learning and
correspondence prediction together in an end-to-end manner. \textcolor{black}{The proposed feature matching method can provide an efficient solution to the visual matching problem in multimedia tasks. }
\item Extensive experiments on three widely used feature matching datasets demonstrate the effectiveness and efficiency of our proposed method.
\end{itemize}

The organization of this paper is described as follows. In Section~\ref{relatedworks}, we give a review of the works most related to our work. In Section~\ref{method}, a comprehensive description of our proposed AMatFormer is given, with a focus on initial feature extraction, anchor matching transformer, and matching prediction. After that, we conduct extensive experiments on multiple benchmark datasets to validate the effectiveness of our model and report the detailed results in Section~\ref{experiments}. Finally, we give the summarization and the future works in Section~\ref{conclusion}.

\section{Related Work} \label{relatedworks}

In this section, we give an introduction to Feature Matching, Graph Matching, Self-Attention and Transformer. 

\subsection{Feature Matching}
The purpose of feature matching is to find the corresponding relationship between the given two features. Usually, the SIFT~\cite{2012Three}, SuperPoint~\cite{2017SuperPoint}, and ContextDesc~\cite{2019ContextDesc} are adopted as the feature extractor. Then, the key research point is to design models to enhance the feature representation and filter out the mismatched point pairs~\cite{ 2017GMS, 2009SCRAMSAC} for final feature correspondences. 
In recent years, many deep learning based feature matching models are proposed. 
For example, 
Detone et al.~\cite{2017SuperPoint} propose a supervised learning training strategy to adaptively detect feature points which inspires many following-up research works. 
Luo et al.~\cite{2019ContextDesc} present a unified learnable framework for enhancing local feature descriptors through cross-modality context information.
Sun et al.~\cite{Sun_2021_CVPR} propose a local image feature matching method based on pixel-level dense matching at coarse-level and then  accurate matching at fine-level.
Jiang et al.~\cite{Jiang_2021_ICCV} propose a Transformer based image matching framework that captures local and global priors. 
Zhao et al.~\cite{9386055} demonstrate that the spatial distribution prior of key points can be assisted by the motion estimation from IMU (inertial measurement unit).  
Pan et al.~\cite{9504452} propose the neighborhood topology consistent descriptor (TCDesc) for image matching. 
Fu et al.~\cite{9905641} propose a Scale-Difference-Aware Image Matching method (SDAIM) to address the issue of large-scale changes in images.  
Fu et al.~\cite{Ma2022} attempt to remove the mismatches by proposing the motion-consistency driven matching (MCDM) scheme. 
Ma et al.~\cite{MA2022196} eliminate mismatched features using their neighborhood manifold representation consensus (NMRC) feature matching method.

Recently, there are also some methods to transform feature matching into graph matching to solve the problem. To be specific, Sarlin et al.~\cite{2020SuperGlue} propose a neural network that matches the local features of the two graphs via an attention mechanism. Chen et al.~\cite{chen2021sgmnet} propose a highly sparse graph neural network, which makes the network more concise and efficient by introducing a seeding mechanism into their feature matching framework. Although these works achieve good performance on existing benchmark datasets, however, most of them directly conduct the feature learning and matching procedure on the whole graph points, which limits their efficiency severely. In contrast, our proposed AMatFormer conducts feature matching only on selected node subset, which enables us to achieve effective and efficient feature matching.

\subsection{Graph Matching} 
The goal of graph matching is to find the corresponding relationship between the nodes of two input graphs. It is usually treated as a combinatorial optimization problem and solved using learning-based methods. 
Zanfir et al.~\cite{zanfir2018deep} propose an end-to-end graph matching framework, which uses the existing CNN to extract the deep features from images to construct a similarity matrix and obtain the matching results in an end-to-end manner. Wang et al.~\cite{wang2020combinatorial} propose to employ graph convolutional networks to transform graph matching into a linear assignment problem and learn reliable node embeddings for graph matching tasks. Jiang et al.~\cite{jiang2019glmnet} combine graph learning and graph matching in an end-to-end unified framework that can learn more reliable graph representation. In order to obtain a more consistent representation of nodes, Fey et al.~\cite{fey2020deep} integrate neighborhood consensus into the graph matching framework. Rol{\'\i}nek et al.~\cite{rolinek2020deep} propose an end-to-end trainable structure for deep graph matching problems by developing an optimized combinatorial solver. Liu et al.~\cite{liu2021stochastic} propose a stochastic iterative graph matching model (SIGMA) to explore more possible matches by defining the matching distribution of graph pairs. 
In order to generate consistent node embeddings for source and target graphs, Jiang et al.~\cite{jiang2021gamnet} propose GAMnet to integrate graph adversarial learning and graph matching together into a unified end-to-end network. Liu et al.~\cite{LIU2023109059} propose an attention based framework (GLAM) to enhance the representation of features through self-attention and cross-attention. More recently, 
Jiang et al.~\cite{Jiang_2022_CVPR} formulate graph matching as an integer linear programming (ILP) problem and propose a new graph context-aware network (GCAN) to address graph matching with scale variations. Ren et al.~\cite{Ren_2022_CVPR} propose Appearance and Structure Aware Robust Graph Matching (ASAR-GM) to defend against adversarial attacks and improve the robustness of graph matching. Hou et al.~\cite{HOU2023109035} reformulate the hypergraph matching problem as a noncooperative multi-player game to obtain consistent feature representations, which can thus eliminate the incorrect matches effectively. 
Different from previous works, our proposed AMatFormer is developed based on the Transformer network, which models the relations between different feature points from both source and target domains simultaneously and efficiently.

\subsection{Self-Attention and Transformer} 
Transformer is first proposed to address the problem of natural language processing (NLP) based on the attention mechanism. Later, such network architecture is extended into many other applications~\cite{zhao2023transformer, wang2022hardvs, wang2023mmpts}, for example, Dosovitskiy et al.~\cite{dosovitskiy2021an} propose the visual Transformer network architecture ViT which has been widely used in the computer vision (CV) community. Liu et al.~\cite{Liu_2021_ICCV} propose the Swin-Transformer network which is a hierarchical Transformer developed based on ViT. Touvron et al.~\cite{touvron2021training} propose a teacher-student strategy based on  distillation to enable students to better learn from teachers through data-efficient image Transformers. The knowledge distillation method is proposed by Caron et al.~\cite{Caron_2021_ICCV} to use the smaller patch and vision Transformer together to simplify the self-monitoring training. Yuan et al.~\cite{Yuan_2021_ICCV} propose a new tokens-to-token vision Transformer, termed T2T-ViT, which recursively aggregates adjacent tokens into one token to structurize the image progressively. 
Wu et al.~\cite{Wu_2021_ICCV} propose a convolution visual Transformer (CVT) that introduces convolution to improve the performance and efficiency of visual Transformer. Touvron et al.~\cite{Touvron_2021_ICCV} propose a deeper and quick converging Transformer based on LayerScale for accurate visual recognition. 
Zhai et al.~\cite{Zhai_2022_CVPR} improve the memory consumption and recognition performance of the ViT model by using a scaling technique. Cao et al.~\cite{9784827} propose a Vision-enhanced and Consensus-aware Transformer (VCT), which uses visual information and consensus knowledge to learn image representations.

In addition, Transformers have also been widely used in many other fields. 
For example, Dong et al.~\cite{Dong_2022_CVPR} develop a Transformer based cross-shaped window self-attention mechanism for visual representation learning. 
Cao et al.~\cite{9864614} first propose the VDTR that uses a Transformer to model space and time domain for deblurring. 
Arnab et al.~\cite{Arnab_2021_ICCV} design a Transformer based video classification model. Ranftl et al.~\cite{Ranftl_2021_ICCV} propose a dense prediction transformer to solve the dense prediction task by using a visual Transformer instead of the convolutional network. Liang et al.~\cite{Liang_2021_ICCV} propose the SwinIR based on a Transformer network to restore high-quality images. 
Chen et al.~\cite{Chen_2021_ICCV} propose a dual-branch vision Transformer to learn multi-scale features. 
Tang et al.~\cite{tang2022transsft} exploit the spatial Gaussian prior and frequency information to help improve the visual representation of Transformer networks for visual tracking. 
Chen et al.~\cite{Chen_2021_CVPR} propose a new feature fusion network to improve the relevance in the tracking field, which enhances feature representation through self-attention and cross-attention mechanisms. 
Inspired by these works, in this paper, we propose a novel AMatFormer framework for the feature matching task, which conducts both self- and cross-attention for the selected graph nodes for efficient feature matching.

\section{Our Proposed Approach} \label{method}
In this section, we will first give an overview of our proposed AMatFormer. Then, we will introduce each module in a more detailed way. As shown in Fig.~\ref{framework}, our model contains three main parts, i.e., Feature Extraction, Anchor Matching Transformer, Metric Learning, and Matching Prediction. We will introduce these modules in the following paragraphs, respectively.

\subsection{Overview} 
Given two source images $I^s, I^t$, we first extract the features and integrate the position embedding information as the initial feature representations. Then, we propose the Anchor Selection Module to select matched anchor features from the source and target feature sets. The Anchor Self-attention Module and Anchor Cross-attention Module are proposed to enhance matched anchor features by fusing their self-context information and cross-image information. After that, the Anchor Primary Attention Module is used to fuse the enhanced anchor feature points and the feature points of the original source and target images. The FFN (Feed Forward Network) is adopted to embed the obtained features into a shared feature representation space. Finally, we utilize effective metric learning to learn the similarities between the source and the target feature points, and iteratively calculate the matching results through the Sinkhorn assignment algorithm.

\subsection{Initial Feature Extraction}
Given the source and target images $I^s, I^t$, we first extract the feature points for them respectively. For each key point, we then extract an initial visual descriptor, such as SIFT, CNN, etc. For deep feature matching approaches, CNN feature extraction is also commonly used.
Let $F^s = \{\textbf{f}^s_1, \textbf{f}^s_2, \cdots, \textbf{f}^s_i, \cdots, \textbf{f}^s_n\}$ and $F^t = \{\textbf{f}^t_1, \textbf{f}^t_2, \cdots, \textbf{f}^t_j, \cdots, \textbf{f}^t_m\}$ denote the
collections of feature points in image $I^s, I^t$ respectively.
In addition to the initial visual descriptor of each point $i$, we also map the position of each feature point into the high-dimensional feature space by using Multi-Layer Perceptron (MLP) and combine visual and position feature descriptors together as the feature of each point, as suggested in work~\cite{2020SuperGlue}. More in detail, two kinds of features are utilized in our matching framework, i.e., the traditional manually designed feature \emph{RootSIFT}~\cite{2012Three} and deep learning feature \emph{SuperPoint}~\cite{2017SuperPoint} and about 1000 points are extracted independently from the source and target images, whose dimensions are all 128-D. 

Based on the above feature extraction, one can directly conduct feature interactive learning and matching process. However, the discriminativity of the above local CNN descriptor is generally limited which may lead to mismatches. For the feature matching problem, it is necessary to incorporate a feature enhancement module to obtain more compact and discriminative feature representations for each feature point.
Inspired by recent self-/cross-attention based feature enhancement methods~\cite{2020SuperGlue,chen2021sgmnet}, in the following section, we propose a novel Anchor Matching Transformer (AMatFormer)
for feature enhancement learning, as introduced below.

\begin{figure*}[t]
\centering
\includegraphics[width=1\textwidth]{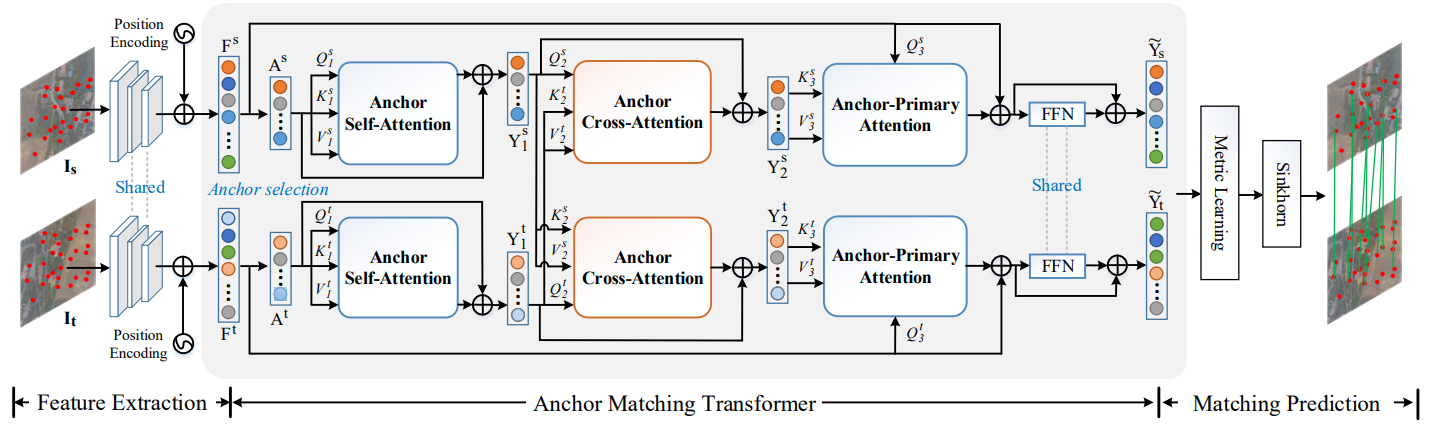} 

\caption{An overview of our proposed efficient feature matching framework. It contains three main modules, i.e., the feature extraction, AMatFormer, and matching prediction module. Note that, our proposed AMatFormer only conduct feature relation learning and matching on a subset of feature points, which decrease the computational cost significantly.
}

\label{framework}
\end{figure*}

\subsection{Anchor Matching Transformer}
\textcolor{black}{As shown in Fig.~\ref{framework}}, our proposed AMatFormer contains the following four modules, i.e., Anchor  Selection, Anchor Self-Attention, Anchor Cross-Attention, and Anchor-Primary Attention module.

\subsubsection{Anchor  Selection Module}
Given the initial feature points obtained from the feature extraction phase, we first need to select a small number of but representative anchor points. Specifically, the nearest neighbour matching is adopted for the initial matching of source and target features. Following existing work~\cite{2004Distinctive}, for each feature point, we treat the ratio of first and second nearest neighbour key points as the reliability score and then choose the top-$k$ matching pairs as the anchor points. The obtained matched anchor features for source and target images are denoted as
$A^s = \{\textbf{a}^s_1, \textbf{a}^s_2, \cdots, \textbf{a}^s_k \}\in \mathbb{R}^{k\times c}$ and $A^t = \{\textbf{a}^t_1, \textbf{a}^t_2, \cdots, \textbf{a}^t_k \}\in \mathbb{R}^{k\times c}$ respectively.

\subsubsection{Anchor Self-Attention Module}
Once we obtained the anchor feature points $A^s$ and $A^t$, we firstly conduct self-attention for each stream independently to capture the  relationships of anchors  within each image, i.e., each anchor learns its context-aware representation by
aggregating the information of other anchors in each image~\cite{chen2021sgmnet}.
Let's take the source anchors  $A^s$ as an example to demonstrate this procedure.
The target anchor points $A^t$ are processed in the similar way. Specifically, we first adopt three linear projection layers to transform  $A^s$ into query ${Q}_1^s$, key ${K}_1^s$, and value ${V}_1^s$ respectively. Then, we compute the correlation/affinity matrix $\mathrm{SAttn}_{s \leftrightarrow s}({Q}_1^s, {K}_1^s)$ between different anchor feature points as follows,
\begin{flalign}
& \mathrm{SAttn}_{s \leftrightarrow s}({Q}_1^s, {K}_1^s) =  Softmax(\frac{{Q}_1^s  {K}_1^{sT}}{\sqrt{c}})
\end{flalign}
where $c$ denotes feature dimension.
Then, each anchor updates its representation by aggregating the messages from the other anchors, followed by layer normalization and residual operation as
\begin{flalign}
&Y^s_1 = A^s + LP(\mathrm{SAttn}_{s \leftrightarrow s}({Q}_1^s, {K}_1^s) {V}_1^s)
\end{flalign}
where  $LP (\cdot)$ refers to the linear projection.

Similarly, we can also obtain the correlation matrix $\mathrm{SAttn}_{t \leftrightarrow t}({Q}_1^t, {K}_1^t)$ for the target anchor points as
\begin{flalign}
& \mathrm{SAttn}_{t \leftrightarrow t}({Q}_1^t, {K}_1^t) = Softmax(\frac{{Q}_1^t  {K}_1^{tT}}{\sqrt{c}})
\end{flalign}
The context enhanced representations for target anchor points can similarly be obtained via the attention aggregation and layer normalization and residual connection operation as
\begin{flalign}
&Y^t_1 = A^t + LP(\mathrm{SAttn}_{t \leftrightarrow t}({Q}_1^t, {K}_1^t) {V}_1^t)
\end{flalign}
where $LP (\cdot)$ refers to linear projection.

\subsubsection{Anchor Cross-Attention Module}
For the matching problem, it is necessary to conduct cross learning on both source and target images to achieve information communication on two images~\cite{2020SuperGlue,chen2021sgmnet,liu2021joint}. Therefore, after obtaining the $Y^s_1$ and $Y^t_1$ via the above self-attention module, we also design the cross-attention module to boost the interactive learning between source and target image anchor features.

To be specific, to achieve interactive information propagation from target to source points, we first transform  $Y^s_1$, $Y^t_1$ and $Y^t_1$ into input query ${Q}_2^s$, key ${K}_2^t$ and values ${V}_2^t$ respectively by using three different linear transformations. Then, the cross-attention weights  $\mathrm{CAttn}_{t \rightarrow s}({Q}_2^s, {K}_2^t)$ from source to target anchor points can be computed as
\begin{flalign}
&\mathrm{CAttn}_{t \rightarrow s}({Q}_2^s, {K}_2^t) = Softmax(\frac{{Q}_2^s  {K}_2^{tT}}{\sqrt{c}})
\end{flalign}
Using this cross affinity,
the anchor features $Y^s_1$ can be enhanced by aggregating the information from target anchors.
This can be achieved  by
using the message passing procedure as follows, 
\begin{flalign}
&Y^s_2 = Y^s_1 + LP(\mathrm{CAttn}_{t \rightarrow s}({Q}_2^s, {K}_2^t) {V}_2^t)
\end{flalign}
Similarly, we can compute the cross-attention affinity matrix  $\mathrm{CAttn}_{s \rightarrow t}({Q}_2^t, {K}_2^s)$ from source to target image via the following procedure as
\begin{flalign}
&\mathrm{CAttn}_{s \rightarrow t}({Q}_2^t, {K}_2^s) = Softmax(\frac{{Q}_2^t  {K}_2^{sT}}{\sqrt{c}})
\end{flalign}
Then, the enhanced features can be similarly updated via the message passing procedure as
\begin{flalign}
&Y^t_2 = Y^t_1 + LP(\mathrm{CAttn}_{s \rightarrow t}({Q}_2^t, {K}_2^s) {V}_2^s)
\end{flalign}
where $LP (\cdot)$ refers to the linear projection.

\begin{figure}[t]
\centering
\includegraphics[width=0.4\textwidth]{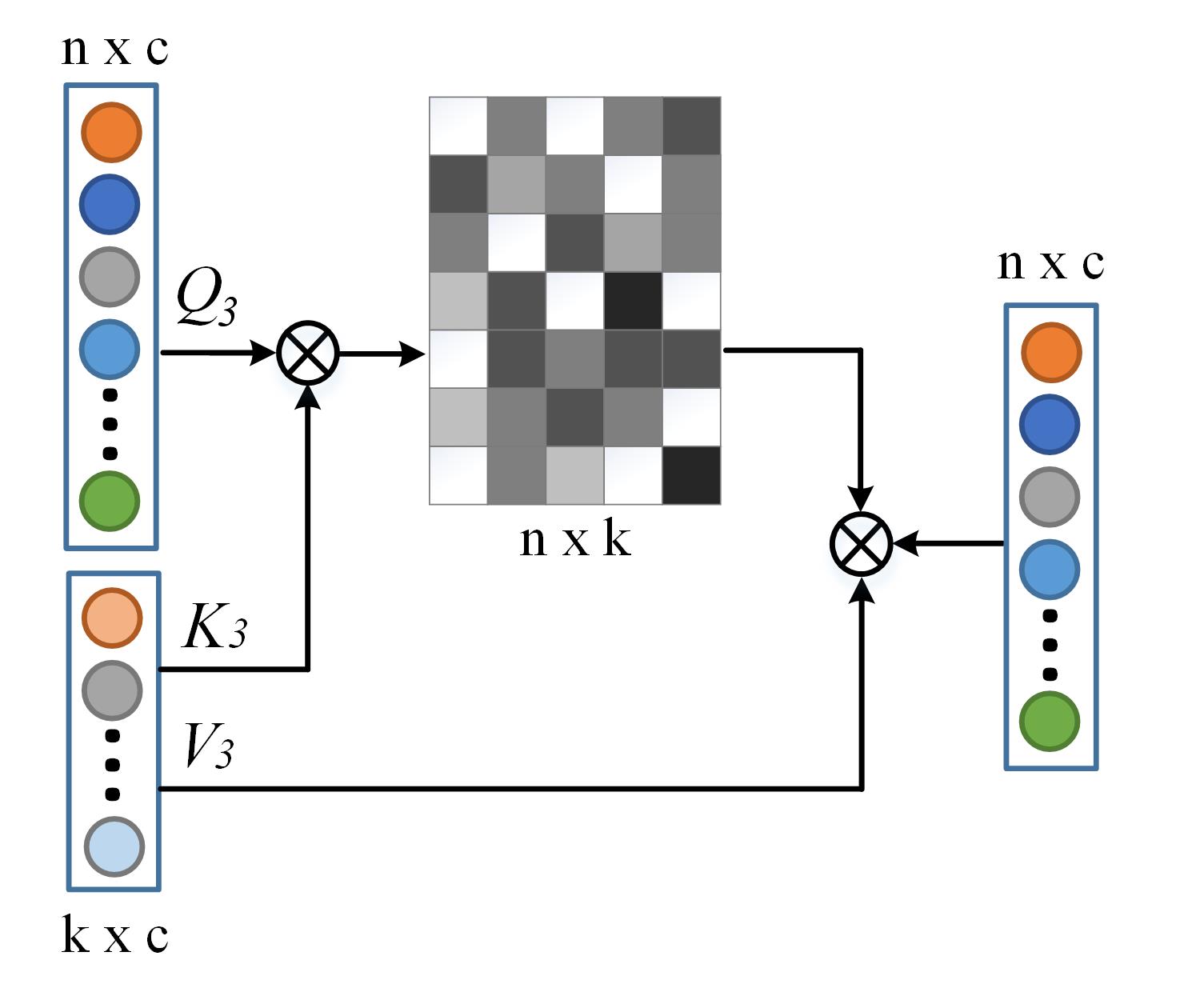} 
\caption{Illustration of Anchor-Primary Attention Module used in our feature matching framework.}
\label{subflow}
\end{figure}

\subsubsection{Anchor-Primary Attention Module}
After the cross interactive learning between source and target anchor points, we leverage them
as bottlenecks to update/learn the representations for the primal features in both images for the subsequent matching procedure, as suggested in works~\cite{chen2021sgmnet,lee2019set}. %
In this work, the anchor-primary attention module is proposed to achieve this goal, as shown in Fig.~\ref{subflow}. 
Specifically, we first transform the above output $Y^s_2$ (Eq.(6)) into key ${K}_3^s$ and value ${V}_3^s$ by using three linear projections respectively and regard the original primary feature points as the query ${Q}_3^s\in \mathbb{R}^{n\times c}$. Then, we compute the attention weights between primary and anchor points in the source image as
\begin{flalign}
&\mathrm{Attn}_{s \rightarrow s}({Q}_3^s, {K}_3^s) = Softmax(\frac{{Q}_3^s  {K}_3^{sT}}{\sqrt{c}})
\end{flalign}
Finally, the  primary features of source image can be updated by using the following message mapping function~\cite{lee2019set} as %
\begin{flalign}
&Y^s_3 =F^s + LP(\mathrm{Attn}_{s \rightarrow s}({Q}_3^s, {K}_3^s) {V}_3^s).
\end{flalign}
where $Y^s_3\in \mathbb{R}^{n\times c}$ denotes the learned representations for all primary feature points in source image. 
Similarly, for the target primary features, we adopt a similar attention scheme to get  $\mathrm{Attn}_{t \rightarrow t}({Q}_3^t, {K}_3^t)$ as
\begin{flalign}
&\mathrm{Attn}_{t \rightarrow t}({Q}_3^t, {K}_3^t) = Softmax(\frac{{Q}_3^t  {K}_3^{tT}}{\sqrt{c}})
\end{flalign}
The attention weights are used to learn the representations for the primary features in target image as
\begin{flalign}
&Y^t_3 = F^t + LP(\mathrm{Attn}_{t \rightarrow t}({Q}_3^t, {K}_3^t) {V}_3^t)
\end{flalign}

In addition, to embed the source and target features into the consistent feature space for feature matching task, we adopt a shared FFN (Feed Forward Network) layer~\cite{2017Attention} to transform  $Y^s_3$ and $Y^t_3$ according to the following formula:
\begin{equation}
\tilde{Y^s} = FFN(LN(Y^s_3),\Theta) + Y^s_3; \tilde{Y^t} = FFN(LN(Y^t_3),\Theta) + Y^t_3
\end{equation}
where $LN (\cdot)$ represents layer normalization and $\Theta$ denotes the parameters of FFN which are shared on both source and target images to obtain consistent feature representations for source and target feature points.

\subsection{Metric Learning and Matching Prediction}
Using
the above AMatFormer model, we obtain the final feature representations $(\tilde{Y}^s \in \mathbb{R}^{n\times c},\tilde{Y}^t \in \mathbb{R}^{m\times c})$ for all feature points in two images.
We use the metric learning method~\cite{wang2019learning} to compute the metric/similarity matrix $S$ between feature points in source and target images as
\begin{align}\label{EQ:layer_cgcn}
S(i,j) = \tilde{Y}^s(i) {W} \tilde{Y}^t(j)^T,
\end{align}
where ${W} \in \mathbb{R}^{c\times c}$ denotes the learnable weight parameters and $S(i,j)$ represents the similarity between  point $i$ of the source image and  point $j$ in the target image.

After obtaining the metric matrix $S$, one can use a linear assignment algorithm to achieve a matching prediction between feature points of two images. We adopt the commonly used linear assignment algorithm Sinkhorn~\cite{sinkhorn1967concerning}, i.e.,
\begin{align}
\tilde{S} = \textrm{Sinkhorn}(S)
\end{align}
Note that, Sinkhorn~\cite{sinkhorn1967concerning} algorithm is differentiable and can be integrated with the above learning modules together in an end-to-end manner.
Let $T$ be the ground-truth  correspondence set between source and target features. Also, let $U^s$ be the set of source image features that do not have the corresponding features in target image $I^t$ and $U^t$ is similarly defined.
Then, the matching prediction loss
 function used in the training phase can be expressed as
\begin{align}
&L_m = \\
&-\big(\sum_{(i,j)\in T}^{} log\tilde{S}_{i,j}+\sum_{i\in U^s}^{} log\tilde{S}_{i,m+1}+\sum_{j\in U^t}^{} log\tilde{S}_{n+1,j}\big)\nonumber
\end{align}
In addition, we also mark the anchor correspondence as matched pairs if their epipolar distance is less than a threshold.
Following SGMnet~\cite{chen2021sgmnet}, we calculate the cross entropy loss of matched and unmatched binary classifications in the $r^{th}$ processing unit and denote it as $L^r$ where $r \in \{1,2\dots R\}$ and $R$ is the number of running units of the proposed AMatFormer module. Therefore, the final total loss can be expressed as:
\begin{align}
Loss = L_m+\alpha \sum\nolimits_{r=1}^{R}L^r
\end{align}
where $\alpha$ is the balanced parameter.

\section{Experiments} \label{experiments}

In this section, we evaluate the effectiveness of the proposed AMatFormer on several benchmark datasets and compare it with some other deep feature matching approaches including 
AdaLAM(4k)\cite{2020Handcrafted}, 
OANet\cite{2020Learning}, SuperGlue\cite{2020SuperGlue} and SGMNet\cite{chen2021sgmnet}. Both SuperGlue\cite{2020SuperGlue} and SGMNet\cite{chen2021sgmnet} adopt self-attention learning for feature matching tasks and thus are most related with our AMatFormer work.

\subsection{Datasets and Evaluation Metric}
In our experiments, our model is trained on \textbf{GL3D}~\cite{shen2018matchable} dataset, and is tested on three benchmark datasets, including \textbf{Scannet}~\cite{2017BundleFusion}, \textbf{Fm-bench}~\cite{2019An} and \textbf{YFCC100M}~\cite{2016YFCC100M}. 
In our experiments, we use RANSAC for post-processing of Scannet~\cite{2017BundleFusion} and YFCC100M~\cite{2016YFCC100M} datasets to remove outliers. 
A brief introduction to these datasets can be found in the following paragraphs. 

$\bullet$ \textbf{GL3D} dataset~\cite{shen2018matchable} contains a large number of 3D geometric structures. 
The dataset contains multi-scale and multi-angle images covering cities, rural areas, scenic spots, landmarks and small objects.

$\bullet$ \textbf{Scannet}~\cite{2017BundleFusion} contains 21 categories of objects which covers 1513 collected scene data. 
Following the experimental criteria of Superglue~\cite{2020SuperGlue}, we selected 1500 pairs of data in the testing set to evaluate our model.

$\bullet$ \textbf{Fm-bench}~\cite{2019An} consists of four subsets in different scenarios, including KITTI~\cite{2012Are}, TUM~\cite{2012A}, T\&T~\cite{2017Tanks} and CPC~\cite{2014Robust}. In this dataset, we selected 4000 pairs for testing. KITTI~\cite{2012Are} is a public dataset obtained by vehicle data collection for the actual traffic scene. 
TUM~\cite{2012A} consists of 39 sequences recorded in the indoor environment. Tanks and Temples (T\&T) ~\cite{2017Tanks} and CPC~\cite{2014Robust} are used for reconstruction tasks.

$\bullet$ \textbf{YFCC100M}~\cite{2016YFCC100M} is an image database based on Yahoo Flickr, which is composed of 100 million pieces of media data generated from 2004 to 2014, including 99.2 million pieces of photo data and 800000 pieces of video data. We also take 4000 pairs in this dataset for testing.

We use the evaluation metric \textbf{AUC}, \textbf{M.S.} (Mean matching Score), and \textbf{Precision} to evaluate the effectiveness of different methods.
\textbf{AUC} is used to measure the angle difference between the ground truth value of rotation and translation and the estimated vector.
\textbf{M.S.} represents the ratio of correct matching and total feature points.
\textbf{Precision} represents the average accuracy of the matching results.
For FM-Bench dataset~\cite{2019An}, \textbf{Recall} and \textbf{Corrs (-m)} are used for the evaluation where
\textbf{Corrs (-m)} represents the average number of correct correspondences after/before RANSAC.

\subsection{Implementation Details}
Following previous works~\cite{chen2021sgmnet}, we first train our model on GL3D dataset~\cite{shen2018matchable}, then test it on three datasets including Scannet~\cite{2017BundleFusion}, Fm-bench~\cite{2019An} and YFCC100M~\cite{2016YFCC100M} datasets. 
The initial learning rate is set as 1e-4, and 900K iterations are trained. We use learning rate decay from 300K iteration to 900K iteration. The batchsize is set to 16 during the training phase. The number $k$ of anchor points in each image is set to 128, as suggested in work~\cite{chen2021sgmnet}.
The number of iterations for sinkhorn is 10. 
\textcolor{black}{Following previous experiments~\cite{chen2021sgmnet}, we set the parameter $\alpha$ as 250 to balance two loss functions. We set three layers for the self-attention and cross-attention modules, respectively.} The experiments are conducted on a server with GPU Tesla P100-PCIE-16GB for both training and testing. Our code is developed based on Python 3.7 and PyTorch 1.8~\cite{paszke2019pytorch}.

\subsection{Comparison Results}
\textcolor{black}{In this section, we evaluate the feasibility of our method and compare it with some other recent competitive models on three image matching datasets. 
The results of all compared methods are reported either  in the original papers or  in work~\cite{chen2021sgmnet}. 
Here, we use them directly in our experiments. 
SGMNet~\cite{chen2021sgmnet} has been shown obviously outperforming previous methods. Therefore, in our experiments, we mainly compare our method with SGMnet, SuperGlue and OANet. }





\begin{table}
\caption{Experimental results on the Scannet dataset. The best results are marked as \textbf{bold}.} 
\label{b2}
\begin{tabular}{c|c|ccc|c|c}
\hline
\hline
\multirow{2}{*}{\textbf{Feature}} &
\multirow{2}{*}{\textbf{Method}} &
\multicolumn{3}{c|}{\textbf{AUC}}&
\multirow{2}{*}{\textbf{M.S.}}&
\multirow{2}{*}{\textbf{Prec.}} \\
\cline{3-5}
& &@$5^{\circ}$ &@$10^{\circ}$ &@$20^{\circ}$ & & \\
\hline
\multirow{4}{*}{\textbf{RootSIFT}} 
 &AdaLAM(4k)  &8.24  &18.57  &31.01  &3.10  & \textbf{47.59}\\
 &OANet  &10.71  &23.10  &37.42  &3.20  &36.93\\
 &SuperGlue &\textbf{13.12}  &27.99  &43.92  &8.50  &42.53\\
 &SGMNet    &12.82 &27.92  &44.55  &8.79  &45.55\\
 &AMatFormer &12.75 &\textbf{28.50} &\textbf{45.11} &\textbf{9.41} &44.26 \\ 
\hline
\multirow{5}{*}{\textbf{SuperPoint}} 
 &AdaLAM(4k)  &6.72  &15.82  &27.37  &13.19  &44.22\\
 &OANet  &10.04  &25.09  &38.01  &10.56  &44.61\\
 &SuperGlue   &13.95  &29.48  &46.07  &15.82  &44.18\\
 &SuperGlue*  &16.19  &\textbf{33.82}  &\textbf{51.86}  &\textbf{18.50}  &47.32\\
 &SGMNet   &15.40  &32.06  &48.32   &16.97  &\textbf{48.01}\\
&AMatFormer &\textbf{16.25} &32.22 &48.26 &17.43 &47.08 \\
\hline
\end{tabular}
\end{table}

\begin{table*}[!htp]
\caption{Experimental results on the FM-Bench. The best results are marked as \textbf{bold}.} \label{b3}
\begin{center}
%
%
\scalebox{1.1}{
\begin{tabular}{c|c|cc|cc|cc|cc}
\hline
\hline
\multirow{2}{*}{\textbf{Feature}} &
\multirow{2}{*}{\textbf{Method}} &
\multicolumn{2}{c|}{\textbf{CPC}}&
\multicolumn{2}{c|}{\textbf{T\&T}}&
\multicolumn{2}{c|}{\textbf{TUM}}&
 \multicolumn{2}{c}{\textbf{KITTI}}\\
\cline{3-10}
   & &\%Recall &\#Corrs(-m)&\%Recall & \#Corrs(-m)&\%Recall & \#Corrs(-m)&\%Recall &\#Corrs(-m)\\
   \hline
\multirow{5}{*}{\textbf{RootSIFT}}
 &OANet &58.6&119(167)&84.7&219(306)&62.3&454(396)&89.0&773(854)\\
 &SuperGlue  &61.1&218(466)&\textbf{86.8}&382(767)&65.9&655(1037)&91.0&\textbf{1261(1746)}\\
 &SGMNet &\textbf{62.0}&248(524)&85.9&397(789)&\textbf{66.6}&704(1132)&\textbf{91.2}&1097(1506)\\
  &AMatFormer &60.8&\textbf{274(724)}&85.6&\textbf{447(983)}&64.2&\textbf{895(1602)}&91.0&1148(1715)\\

\hline
\multirow{5}{*}{\textbf{SuperPoint}}
 &OANet &62.9&186(343)&91.2&280(477)&61.4&332(473)&82.2&482(736)\\
 &SuperGlue  &68.8&287(719)&92.9&414(987)&59.1&512(1038)&88.7&\textbf{957}(1777)\\
 &SuperGlue* &\textbf{76.7}&302(712)&\textbf{96.6}&431(985)&59.0&121(177)&\textbf{89.5}&354(526)\\
 &SGMNet &70.3&327 (829) &93.2&450(1098)&\textbf{65.8}& 666 (1315)&86.3&954(1851)\\
 &AMatFormer&66.4&\textbf{355(947)}&93.5&\textbf{477(1189)}&59.8&\textbf{839(2010)}&88.4&933(\textbf{1991})\\
\hline
\end{tabular}}
\end{center}
\end{table*}

\textbf{Results on Scannet Dataset. }
As shown in Table~\ref{b2}, when the RootSIFT feature is used for evaluation, we can find that our baseline method SGMNet achieves $12.82, 27.92, 44.55$ on the $AUC@5^{\circ}, AUC@10^{\circ}, AUC@20^{\circ}$,  $8.79, 45.55$ on the M.S. and Precision metric, respectively. In contrast, our proposed AMatFormer model attains $12.75, 28.50, 45.11$ on the AUC metric, and $9.41, 44.26$  on the M.S. and Precision evaluation metric, respectively.
It is easy to find that our results are better than the SGMNet on the $AUC@10^{\circ}, AUC@20^{\circ}$, and M.S., and comparable on the rest of evaluation metrics.
\textcolor{black}{In the table, SuperGlue* represents the results obtained from the orginal paper and SuperGlue denotes the results reproduced in work~\cite{chen2021sgmnet} which is conducted on the same experimental setting with our method.} 
%
%
Compared with other  methods, such as SuperGlue and OANet, our experimental results are also better than theirs on most of the evaluation metrics.
These experimental results validate the effectiveness of our model.

\textbf{Results on FM-Bench Dataset. }
As shown in Table~\ref{b3}, the FM-Bench dataset contains four subsets including CPC, T$\&$T, TUM and KITTI. Following SGMNet~\cite{chen2021sgmnet}, we first use RANSAC to process the dataset. Then, the fundamental matrix of each evaluation pair ($i, j$) is estimated. Finally, the difference between the estimated fundamental matrix and ground truth is obtained by using the normalized symmetric epipolar distance (SGD)~\cite{1998Determining}. If the difference between the normalized SGD of estimated value and ground truth is lower than the pre-defined threshold, the estimation is deemed correct. Based on the experimental results reported in Table~\ref{b3}, we can find that our model is slightly better than SGMNet~\cite{chen2021sgmnet}. Specifically, we achieve the best Corrs(-m) results on the CPC, T$\&$T, TUM, i.e., 274 (724), 447 (983), 895 (1602), when the RootSIFT is adopted. 
For the SuperPoint feature based results, we can draw similar conclusions on this benchmark dataset.

\textbf{Results on YFCC100M Dataset. }
As shown in Table~\ref{resultsYFCC100M} in detail, we can note that our baseline SGMNet achieves $62.72, 72.52, 81.48, 17.08, 86.08$ on $AUC@5^{\circ}$, $AUC@10^{\circ}$, $AUC@20^{\circ}$, M.S. and Precision metrics, respectively, when the RootSIFT feature is adopted. Our method attains $63.83, 73.26, 81.98, 19.17, 83.85$ on these five evaluation metrics, which are better than the baseline approach on the $AUC@5^{\circ}$ and M.S. criterion.  For the SuperPoint features, our proposed AMatFormer also beats the baseline on four metrics. It is worthy noting that our results are better than the compared methods on most of the evaluation metrics.

Overall, we can draw the conclusion that our AMatFormer is slightly better than baseline SGMNet on most of the evaluation metrics, but significantly better than most of the other compared models. These experiments fully validate the effectiveness of our proposed method for the feature matching tasks. Moreover, our AMatFormer performs obviously faster than baseline SGMNet, as discussed below.


\begin{table}
\caption{Experimental results on the YFCC100M dataset. The best results are marked as \textbf{bold}.} \label{resultsYFCC100M}
\begin{center}
%
%
\begin{tabular}{c|c|ccc|c|c}
\hline
\hline
\multirow{2}{*}{\textbf{Feature}} &
\multirow{2}{*}{\textbf{Method}} &
\multicolumn{3}{c|}{\textbf{AUC}}&
\multirow{2}{*}{\textbf{M.S.}}&
\multirow{2}{*}{\textbf{Prec.}} \\
\cline{3-5}
   & & @$5^{\circ}$&@$10^{\circ}$&@$20^{\circ}$& & \\
\hline
\multirow{5}{*}{\textbf{RootSIFT}}
 &AdaLAM(4k)&57.78&58.76&68.58&8.23&29.79\\
 &OANet&58.00&67.80&77.46&5.84&81.80\\
 &SuperGlue* &59.25&70.38&80.44&-&-\\
 &SuperGlue &63.82&\textbf{73.33}&\textbf{82.26}&16.59&81.08 \\
 &SGMNet &62.72 &72.52&81.48&17.08&\textbf{86.08}\\
  &AMatFormer&\textbf{63.83} & 73.26 & 81.98 & \textbf{19.17}&83.85  \\ 
\hline
\multirow{5}{*}{\textbf{SuperPoint}}
 &AdaLAM(4k) &40.20&49.03&59.11&10.17&72.57\\
 &OANet&48.80&59.06&70.02&12.48&71.95\\
 &SuperGlue* &\textbf{67.10}&\textbf{76.18}&\textbf{84.37}&21.58&\textbf{88.64}\\
 &SuperGlue &60.37&70.51&80.00&19.47&78.74\\
 &SGMNet &61.22 &71.02&80.45&22.36&85.44\\
 &AMatFormer&62.20&71.86&81.11&\textbf{22.79}&84.82 \\ 
\hline
\end{tabular}
\end{center}
\end{table}

\begin{figure*}
\centering
\includegraphics[width=1\textwidth]{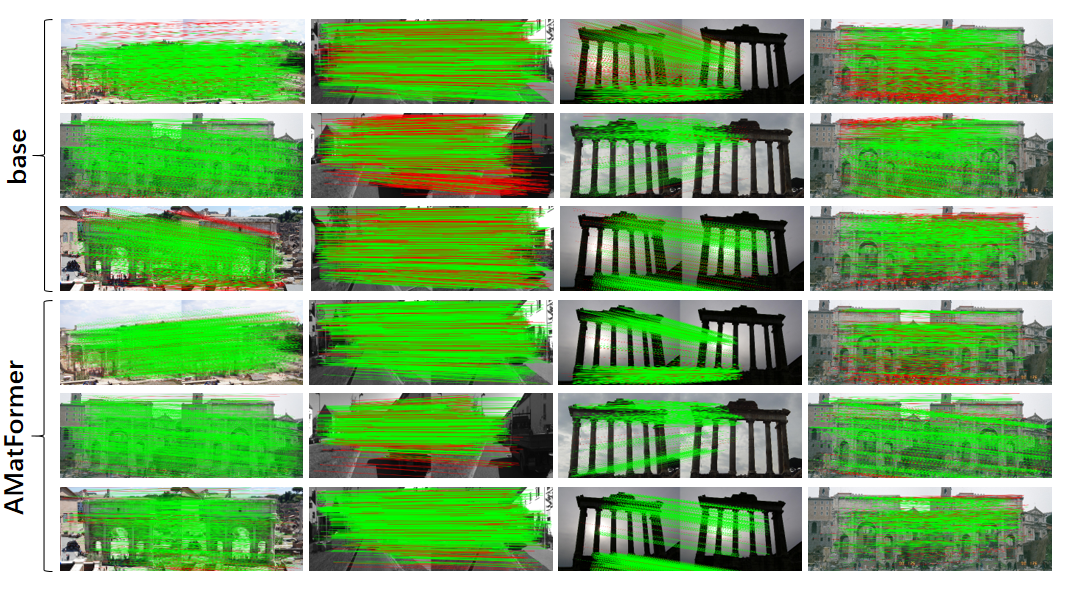} 
\caption{Visualization of feature matching results produced by the baseline SGMNet\cite{chen2021sgmnet} (base in this figure) and our proposed AMatFormer on \textbf{FM-Bench} dataset (RootSIFT feature adopted).}
\label{visualizeFeatureMatchFMbench}
\end{figure*}


\subsection{Efficiency Analysis}

\begin{table}
\centering
\caption{\textbf{Top}: Comparison of Flops, parameters, running time, and peak memory usage;  \textbf{Bottom}: Theoretical analysis of algorithm complexity.} \label{TheoreticalResults}
\small
\begin{tabular}{l|lllll}
\hline
\textbf{Model}     &\textbf{Flops}   &\textbf{Param.} &\textbf{Time}    &\textbf{Memory}  \\
\hline
\textcolor{black}{SuperGlue}    &\textcolor{black}{24.547G}    &\textcolor{black}{13.068M}  &\textcolor{black}{255.72ms}  &\textcolor{black}{874MB}    \\
SGMNet    &11.919G    &7.744M  &118.48ms  &361MB     \\
AMatFormer     &4.747G    &3.875M  &83.97ms  &346MB     \\
\hline
\textcolor{black}{SuperGlue}
&\multicolumn{4}{c}{$\textcolor{black}{3n^2c+4nc^2}$}   \\
SGMNet    &\multicolumn{4}{c}{$2(2nkc+2k^2c+4kc^2+2nc^2)$}    \\
AMatFormer      &\multicolumn{4}{c}{$2(nkc+2k^2c+nc^2)$}     \\
\hline
\end{tabular}
\end{table}

As shown in Table~\ref{TheoreticalResults}, we give the theoretical analysis on the algorithm complexity of our proposed AMatFormer. 
To be specific, 
the main computational complexity involves two main parts. The first part is the message passing operation.  
SGMNet employs attentional pooling, seed filtering and attentional unpooling whose complexity is $\mathcal{O}(nkc), \mathcal{O}(2k^2c)$ and $\mathcal{O}(nkc)$ respectively. The whole complexity for message passing part is $\mathcal{O}(4nkc+4k^2c)$ for both source and target branches. In our AMatFormer, however, it only involves anchor filtering and anchor-primary  attention. Thus, the complexity is $\mathcal{O}(2nkc+4k^2c)$ for two image branches. 
The second part is the feature transformation. %
Since the linear projections to obtain query, key and value have been commonly used in both SGMNet and AMatFormer, we omit them in our analysis. 
Besides, for SGMNet, it adopts a shared one-layer MLP in each attention computation module whose complexity is $\mathcal{O}(2(4kc^2+2nc^2))$. 
In contrast, in our AMatFormer, it only adopts a shared one-layer FFN for the final feature transformation and thus the complexity is only  $\mathcal{O}(2nc^2)$ for two branches. 
Therefore, the total complexity of SGMNet is $\mathcal{O}(2(2nkc+2k^2c+4kc^2+2nc^2))$, meanwhile, ours is only $\mathcal{O}(2(nkc+2k^2c+nc^2))$. It is easy to find that our model can run obviously faster than the baseline approach in theoretical.

In addition to theoretical analysis, we also give a comparison of quantitative results with the baseline model SGMNet on FLOPs, parameters, running time and peak GPU memory usage, as summarized in Table 4. Specifically, the time cost of SGMnet is 118.48 ms, meanwhile, ours are 83.97 ms which is significantly faster than the baseline (up to $29\%$). Our model also contains less parameters than the SGMNet, i.e., Ours 3.875M vs. SGMNet 7.744M. Correspondingly, the FLOPs of the baseline and our model are 11.919G and 4.747G, respectively, which fully demonstrates the efficiency and advantages of our AMatFormer model.

\begin{figure*}
\centering
\includegraphics[width=1\textwidth]{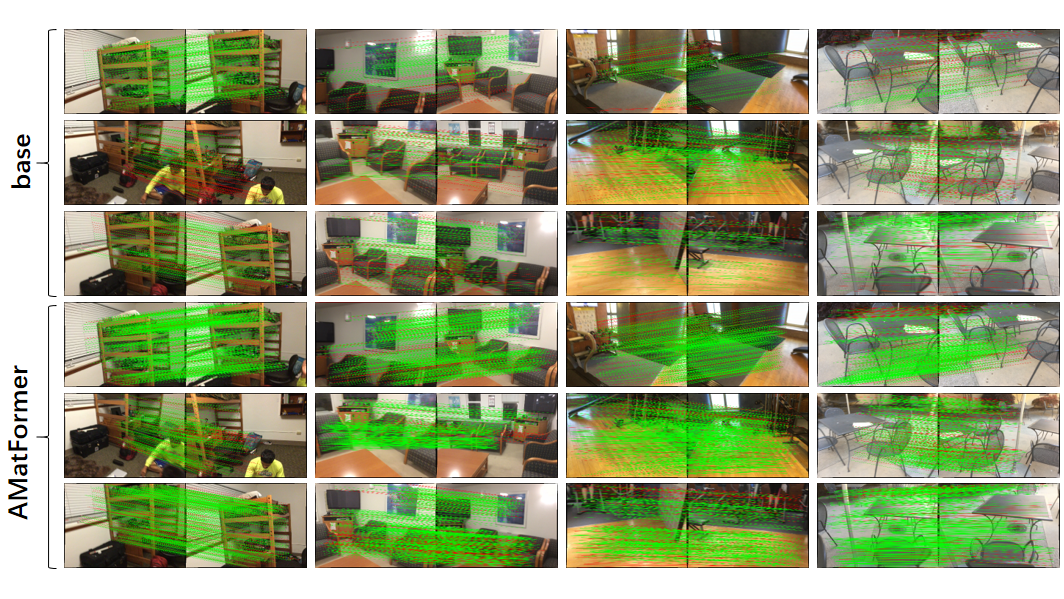} 
\caption{Visualization of feature matching results produced by the baseline SGMNet\cite{chen2021sgmnet} (base in this figure) and our proposed AMatFormer on \textbf{Scannet} dataset (RootSIFT feature adopted).}
\label{visualizeFeatureMatch}
\end{figure*}

\subsection{Ablation Study}
In this section, the FFN module, metric learning algorithms, anchor cross-attention module, and the number of selected anchor features are further analyzed to help readers better understand these modules. 

{\textbf{Effects of FFN layers. }}
The FFN in AMatFormer is an important step to transform the source and target feature points into a shared feature space. In this section, we will test its effectiveness by comparing with the model without FFN layers on the Scannet dataset.
As shown in Table~\ref{ablationFFN}, our model achieves 13.33, 27.78, 44.36, 9.23, 44.61 on $AUC@5^{\circ}, AUC@10^{\circ}, AUC@20^{\circ}$, M.S., Precision metrics, respectively. These results will drop to 12.78, 26.96, 42.74, 8.61, 43.27, if this module is removed.
Obviously, these experiments demonstrate that it is important to transform the source and target features into a shared feature space before conducting feature matching.

\begin{table}[!htp]
\caption{Component analysis of FFN layers on Scannet dataset (RootSIFT is adopted)} \label{ablationFFN}
\begin{center}
\scalebox{1}{
\begin{tabular}{c|c|ccc|c|c}
\hline
\hline
\multirow{2}{*}{\textbf{Method}} &
\multirow{2}{*}{\textbf{FFN}} &
\multicolumn{3}{c|}{\textbf{AUC}}&
\multirow{2}{*}{\textbf{M.S.}}&
\multirow{2}{*}{\textbf{Prec.}} \\
\cline{3-5}
   & & @$5^{\circ}$&@$10^{\circ}$&@$20^{\circ}$& & \\
\hline
\multirow{2}{*}{AMatFormer} &$\times$  &12.78  &26.96  &42.74  &8.61  &43.27  \\
 &\checkmark                         &13.33  &27.78  &44.36  &9.23  &44.61  \\
\hline
\end{tabular}}
\end{center}
\end{table}

\begin{table}[!htp]
\caption{Component analysis of metric learning on YFCC100M dataset (RootSIFT is adopted)} \label{ablationMetric}
\begin{center}
\scalebox{1}{
\begin{tabular}{c|ccc|c|c}
\hline
\hline
\multirow{2}{*}{\textbf{Method}} &
\multicolumn{3}{c|}{\textbf{AUC}}&
\multirow{2}{*}{\textbf{M.S.}}&
\multirow{2}{*}{\textbf{Prec.}} \\
\cline{2-4}
 & @$5^{\circ}$&@$10^{\circ}$&@$20^{\circ}$& & \\
\hline
AMatFormer-Cosine  &62.43 &72.13 &81.38 &19.29 &83.70  \\
AMatFormer-Weight  &63.83 & 73.26 & 81.98 & 19.17&83.85  \\
\hline
\end{tabular}}
\end{center}
\end{table}

\begin{table}[!htp]
\textcolor{black}{
\caption{Component analysis of anchor number on YFCC100M dataset (RootSIFT is adopted)} \label{ablationAnchor}
\begin{center}
\scalebox{1.2}{
\begin{tabular}{c|ccc|c|c}
\hline
\hline
\multirow{2}{*}{\textbf{Method}} &
\multicolumn{3}{c|}{\textbf{AUC}}&
\multirow{2}{*}{\textbf{M.S.}}&
\multirow{2}{*}{\textbf{Prec.}} \\
\cline{2-4}
\cline{2-4}
 & @$5^{\circ}$&@$10^{\circ}$&@$20^{\circ}$& & \\
\hline
8   &56.30 &66.30   &76.43  &14.66 &69.14  \\
16   &58.90 &68.65   &78.28  &17.43 &76.00  \\
32   &59.25 &68.85   &78.41  &\textbf{19.46} &78.44  \\
64   &61.23 &71.38   &80.46  &19.34 &81.92  \\
128  &\textbf{63.83}    & \textbf{73.26}  &\textbf{81.98} &19.17 &\textbf{83.85}  \\
256  &62.93    &72.51   &81.53 &18.22 &83.58  \\
\hline
\end{tabular}}
\end{center}
}
\end{table}

\begin{table}[!htp]
\textcolor{black}{
\caption{Component analysis of anchor selection on YFCC100M dataset (RootSIFT is adopted)} \label{ablationAnchor_1}
\begin{center}
\scalebox{1.1}{
\begin{tabular}{c|ccc|c|c}
\hline
\hline
\multirow{2}{*}{\textbf{Method}} &
\multicolumn{3}{c|}{\textbf{AUC}}&
\multirow{2}{*}{\textbf{M.S.}}&
\multirow{2}{*}{\textbf{Prec.}} \\
\cline{2-4}
\cline{2-4}
 & @$5^{\circ}$&@$10^{\circ}$&@$20^{\circ}$& & \\
\hline
anchor selection  &63.83   & 73.26  &81.98 &19.17 &83.85  \\
all features  &61.68    & 71.75  &80.78 &15.39 &82.94  \\
\hline
\end{tabular}}
\end{center}
}
\end{table}

\begin{table}[!htp]
\caption{Component analysis of anchor cross-attention module on yfcc dataset (RootSIFT is adopted)} \label{ablationcross}    
\begin{center}
\scalebox{0.92}{
\begin{tabular}{c|c|ccc|c|c}
\hline
\hline
\multirow{2}{*}{\textbf{Method}} &
\multirow{2}{*}{\textbf{cross-module}} &
\multicolumn{3}{c|}{\textbf{AUC}}&
\multirow{2}{*}{\textbf{M.S.}}&
\multirow{2}{*}{\textbf{Prec.}} \\
\cline{3-5}
   & & @$5^{\circ}$&@$10^{\circ}$&@$20^{\circ}$& & \\
\hline
\multirow{2}{*}{AMatFormer}&$\times$   &63.02  &72.28  &81.41  &18.95  &82.95  \\
 &\checkmark                         &63.83 & 73.26 & 81.98 & 19.17&83.85  \\
\hline
\end{tabular}}
\end{center}
\end{table}

{\textbf{Effects of different metric learning methods.}}
In this paper, two kinds of metric learning methods are evaluated, including cosine similarity and learning based one (Eq.(14)), termed AMatFormer-Cosine and AMatFormer-Weight, respectively. 
As shown in Table~\ref{ablationMetric}, when the RootSIFT feature is used on YFCC100M dataset, we can find that the experimental results of AMatFormer-Cosine achieve $62.43, 72.13, 81.38, 19.29, 83.70$ on the used five evaluation metrics. In contrast, AMatFormer-Weight obtains $63.83, 73.26, 81.98, 19.17, 83.85$ which are better than the AMatFormer-Cosine on four of five evaluation metrics. Therefore, we can draw the conclusion that our proposed learning based metric learning algorithm works well for the feature matching task.

{\textbf{Analysis on Anchor Cross-Attention. }}
The interactive feature learning between source and target branches plays an important role in our proposed framework. To validate the effectiveness of such a cross-attention module, in this section, we conduct an experiment by removing this module. As shown in Table~\ref{ablationcross}, the performance on the Scannet dataset drops from $63.83, 73.26, 81.98, 19.17, 83.85$ to $63.02, 72.28, 81.41, 18.95, 82.95$. These results demonstrate that the anchor cross-attention scheme indeed models the relations between source and target feature points effectively.

{\textbf{Analysis on the Number of Selected Anchor Features.~}} 
As validated in previous sections, the anchor selection plays an important role in our proposed framework. To further analyze its influence, in this subsection, we test the different numbers of anchor points. We select 8, 16, 32, 64, 128, and 256 anchor points on the YFCC100M dataset for verification and report the results in Table~\ref{ablationAnchor}. It is easy to find that better results can be obtained if 128 anchor points are selected. More anchors do not bring improvements in  matching accuracy. 
\textcolor{black}{
We also conduct an experiment to validate if all the features used can further improve the final performance. As shown in Table~\ref{ablationAnchor_1}, we can find that our proposed anchor model generally obtain the better performance, which fully validated the effectiveness of the proposed anchor based Transformer method for matching problem.
}




\subsection{Visualization} 
To help the readers better understand our proposed AMatFormer for image matching, in this subsection, we give a visualization of our feature matching results on samples from Scannet~\cite{2017BundleFusion} and Fm-bench~\cite{2019An}. As shown in Fig.~\ref{visualizeFeatureMatchFMbench} and Fig.~\ref{visualizeFeatureMatch}, we can find that our proposed AMatFormer can match most of the feature points accurately (highlighted in the green line, the red lines are the results of the false match). 
This intuitively 
demonstrates the 
more effectiveness of the proposed 
Transformer model on feature matching tasks when comparing with baseline method.  



\begin{figure}
\centering
\includegraphics[width=0.5\textwidth]{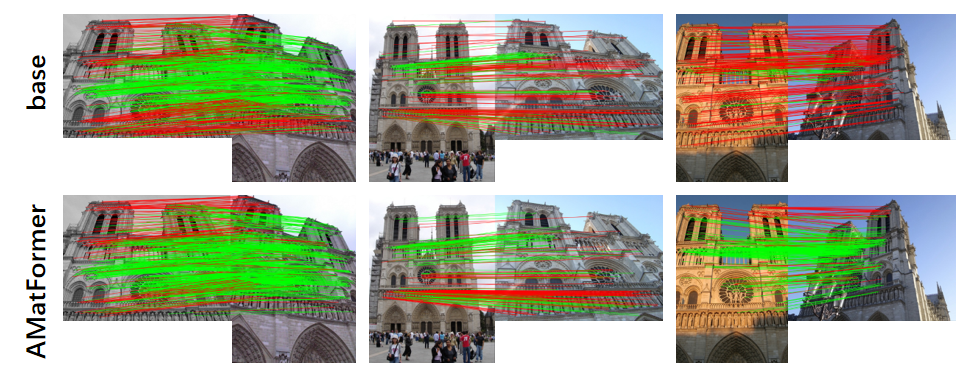} 
\caption{Visualization of  baseline SGMNet\cite{chen2021sgmnet} and our newly proposed AMatFormer on the YFCC100M dataset (RootSIFT feature adopted).}
\label{visualizeyfcc}
\end{figure}

\subsection{Limitation Analysis} 
Although good performance can be achieved by using our proposed AMatFormer on multiple matching benchmark datasets, however, our model still performs poorly when facing challenging situations. As shown in Fig.~\ref{visualizeyfcc}, when significant variations occurred between two buildings, our results are still better than the baseline method. But the overall performance is not satisfactory. Also, our model predicts many incorrect feature matches between the bright and rayless images. In our future works, we will consider learning the illumination invariant features to handle this issue.


\section{Conclusion} \label{conclusion}
In this paper, we propose a novel and efficient Anchor Matching Transformer (AMatFormer) for feature matching task. It simultaneously models the discriminative representations for feature points within each intra-image, and the consensus representations for feature points across inter-images. Our proposed AMatFormer mainly conducts self-/cross-attention on some anchor features instead of all the feature points and thus performs efficiently. Also,  a shared FFN module is used to further project the features of two images into the common domain. On the basis of extensive experiments on three widely used benchmark datasets, we note that our AMatFormer generally achieves the best accuracy meanwhile utilizes  much less time cost. To be specific, it averagely has improved the running efficiency by about $29\%$. We hope our proposed model could provide new insights for the efficient feature matching task. In our future works, we consider designing new metric learning approaches for more efficient feature matching.


\bibliographystyle{IEEEtran}
\bibliography{ref}

\end{document}